\documentclass[conference]{IEEEtran}
\usepackage{amsmath,amssymb}
\usepackage{comment}

\usepackage{placeins}

\usepackage[utf8]{inputenc}
\usepackage[T1]{fontenc}
\usepackage{lmodern}
\usepackage{subcaption}
\usepackage{graphicx}
\usepackage{caption}
\usepackage[numbers]{natbib}   
\usepackage{url}               
\usepackage[hidelinks]{hyperref} 
\usepackage{xurl}               

\usepackage{algorithm}
\usepackage{algpseudocode} 
\usepackage{booktabs}

\title{How Much Can a Few Engine Moves Help? Quantifying Limited Cheating in Chess%
  \thanks{Accepted, IEEE CoG 2026 (IEEE Conference on Games) 2026.}%
}
\author{
  \IEEEauthorblockN{Daniel Keren}
  \IEEEauthorblockA{
    \textit{Department of Computer Science}\\
    \textit{University of Haifa}\\
    Haifa, Israel\\
    \texttt{dkeren@ds.haifa.ac.il}
  }
}

\IEEEoverridecommandlockouts
\IEEEpubid{\makebox[\columnwidth]{979-8-3315-9476-3/26/\$31.00~\copyright2026 IEEE\hfill}\hspace{\columnsep}\makebox[\columnwidth]{ }}

\begin{document}
\maketitle
\IEEEpubidadjcol

\begin{abstract}
Cheating in chess, by using advice from powerful software, has become a major problem, reaching the highest levels. As opposed to the large majority of previous work, which concerned {\em detection} of cheating, here we try to evaluate
the possible gain in performance, obtained by cheating a limited
number of times during a game.
We develop threshold-based and Bellman-style intervention policies, and test them in a controlled engine-vs-engine setting using Stockfish. A judicious choice of 1 or 2 cheats yields average scores of 0.71 and 0.82, respectively, compared to 0.51 with no cheats. We also introduce a fast, engine-free simulator that enables hyperparameter optimization without running games, closely matching the engine-based optimum.

The goal of this work is not to assist cheaters, but to measure the effectiveness of cheating -- which is crucial as part of the effort to contain and detect it.
\end{abstract}
\section{Introduction}
The game of chess is one of the most popular intellectual pastimes world-wide. Millions of people play chess in clubs and tournaments, and far more play for leisure. The popular online platform Chess.com has
about 200 million members; and in the publicly available database
of games in Lichess.org, another popular platform, roughly 100 million games are added every month.

While cheating had always plagued chess, in recent years it had become
a far more pronounced problem, affecting the highest echelon of players, and widely reported in the media. Since cheating is easier in online games, it had reached epidemic proportions, with Chess.com closing about 100,000 accounts per month for violating fair play rules, including a few Grandmasters (the highest rank in chess). This problem is far more pronounced in recent years, as online games have become a major component of chess, with an extensive tournament cycle, in which the world's top players regularly participate. 

High-profile cases such as the six-year FIDE ban of GM Igor Rausis~\citep{FIDE2019RausisDecision} and the 2022 Carlsen--Niemann controversy and subsequent litigation~\citep{Carlsen2022ChessBaseStatement,Chesscom2022NiemannReport,Chesscom2023Settlement} underscore the severity of the problem; for an overview of platform-level fair-play detection, see~\citep{Chesscom2025FairPlayOverview}.

In this paper, we do not study methods to detect cheating, but investigate how effective cheating is, in games between chess engines. Cheating  is defined as the "helper" (a strong engine, that is watching the game) intervening in a few moves, and advising the "cheater" how to play. Since we wish to
investigate the effectiveness of "cheating without
being easily caught", we provide the cheater with
a measure of camouflage, by allowing only a very small number of carefully
chosen intervention moves.

A judicious choice of 1 resp. 2 
cheats in a game leads to an average score of approximately 
0.71 resp. 0.82, vs.\ 0.51 for no cheats (we assume that only the player with the white pieces cheats, and 0.51 reflects the small advantage resulting from White playing the first move. The algorithms presented here apply also to the player with the black pieces).

The cheating method tested here is similar to common real-world cheating
practice: players use a chess engine (powerful software) but try to
minimize the number of assisted moves, in order to avoid detection, since
often, cheating-detection algorithms rely heavily on the presence of
many excellent, difficult-to-find moves. In Section \ref{section:human-human}, we briefly discuss the relevance to human-vs-human cheating,
which we hope to test in the future (however it is logistically difficult, due to the large number of
games required to tune the algorithms and to reach reliable statistics).

Two intervention strategies were tested: one "local", which intervenes if the local gain is beyond an optimized constant, and one "look ahead", which intervenes only if the immediate gain is at least as large as the predicted gain from a future intervention.

\section{Previous work}

There is a huge body of work on computer chess; we briefly survey the history of chess engines, as well as research on detecting cheating in chess, and modeling human behavior.

\subsection{Chess engines (brief overview)}
A complete treatment of chess engines is beyond our scope, but we note two broad strands.
Classical search engines follow the minimax paradigm with alpha--beta pruning, transposition
tables, sophisticated move ordering, and endgame tablebases. The strongest open–source
representative is \emph{Stockfish}, which is widely used as a
reference implementation and evaluation tool; some references (incl. earlier, classical work) are \citep{Shannon1950,KnuthMoore1975,Nalimov2000,
StockfishEngine,StockfishGitHub}.

A second strand integrates learning and search. Google DeepMind's \emph{AlphaZero} demonstrated
self–play reinforcement learning with Monte-Carlo Tree Search, achieving excellent performance 
without handcrafted evaluation features \citep{Silver2018AlphaZero}. In parallel, the neural
evaluation approach \emph{NNUE}  was adapted into classical alpha--beta
frameworks and incorporated into Stockfish, yielding large gains while retaining fast search
\citep{Nasu2018NNUE}. The community–driven \emph{Leela Chess Zero} project provides an open
end–to–end neural alternative, helping standardize training and evaluation practices
\citep{LeelaChessZero}. 

\subsection{Previous work on detecting cheating in chess}

Many studies discuss how to distinguish human play from engine-assisted play.
Early work compared human moves to top engine choices \citep{GuidBratko2006}, though relying solely on engine agreement has known pitfalls \citep{BarnesHernandezCastro2015}. Regan and collaborators proposed Bayesian models of fallible choice that account for position difficulty and player strength \citep{DiFattaHaworthRegan2009,HaworthReganDiFatta2010}, leading to \emph{Intrinsic Chess Ratings} \citep{ReganHaworth2011ICR}. More recent approaches include interpretable human/computer decision criteria \citep{LaarhovenPonukumati2023}, neural classifiers for fair-play detection \citep{IavichKevanishvili2024}, and streak-probability analysis \citep{Rosenthal2024Streaks}. Empirical analyses further show that sparse, well-timed assistance can disproportionately affect outcomes \citep{maharaj2024kramnik}. \cite{haugen2023heuristic} use a player's draw rate as a heuristic indicator
of possible cheating. Chess remains a benchmark for studying evaluation and planning in games \cite{DBLP:journals/tciaig/TaoXZWSLFHZP23,11114281,10333133}, though prior work does not consider selective intervention strategies, as we do here. 

\subsection{Modeling human move behavior in chess}

A complementary line of work models how humans actually choose moves. The Maia project
\citep{McIlroyYoung2020Maia} trains engines to \emph{align with human preferences},
yielding policies that predict human moves at various skill levels. Subsequent work
models individual players’ tendencies \citep{McIlroyYoung2022Individual},
capturing stable, personalized patterns.

For fair-play detection, such human-aligned and per-player models provide a behavioral
baseline: suspicious play can be framed as systematic deviation from (i) general human policies and/or (ii) a player’s own historical policy.
%
\subsection{Other work on assistance/cheating/intervention, and our contribution}
Much of the chess literature on engines and human behavior focuses on \emph{detection} of illicit assistance. A smaller body of research studies how access to assistance changes \emph{performance}. The Maia line of work models human move choice directly and enables counterfactual comparisons between human and engine policies, but it does not quantify the win-rate gain from a fixed number of assisted moves \citep{McIlroyYoung2020Maia,McIlroyYoung2022Individual}. Outside chess, a large-scale analysis of professional Go shows that the advent of superhuman AI improved human decision quality and novelty, using engine-estimated counterfactuals to assess uplift \citep{Shin2023PNAS}. Practitioner-facing write-ups (e.g., Chessable’s limited-assist case study) discuss how players actually deploy engines online \citep{Chessable2024Pt2}.

Our contribution differs in two key points. First, we quantify the \emph{average score uplift} from a strict budget $n$ of oracle (very strong engine) interventions within an engine-vs-engine environment; to our knowledge, this fixed-budget approach has not been reported for chess. Second, our estimator uses logged no-intervention and randomized-intervention data with monotone calibration to produce counterfactuals tailored to threshold policies.

Related paradigms exist in other games. Cooperative games like Hanabi naturally impose limited, discrete \emph{hints}, a close analogue to capped interventions \citep{Bard2019Hanabi}. More broadly, the \emph{action-advising} literature in reinforcement learning studies when and how an oracle/teacher should inject advice under a budget to improve returns—conceptually aligned with our fixed-\(n\) oracle-move interventions \citep{DaSilva2020UncertaintyAdvising}.

Many traditional games alternate turns between players, e.g., tic-tac-toe or Chess. Instead, in bidding games \cite{LLPSU99,AHC19}, players have "move budgets" and in each turn, a bidding determines which player moves. Interesting equivalences have been identified between bidding games and a class of stochastic games called "random-turn games" \cite{PSSW07}, in which the player who moves in each turn is chosen randomly \cite{LLPSU99,AHC19,AJZ21}. 
\section{Setting}
The game of chess is played between a player with the white pieces (denoted 
${\cal \bf{W}}$) who makes the first move, and a player with the
black pieces (denoted ${\cal \bf{B}}$). We do not describe the rules of chess
here; they are readily available. The game can end with 
${\cal \bf{W}}$
winning, with the corresponding result of 1 for the game; a win for ${\cal \bf{B}}$ is assigned the value 0, and a draw, 0.5.

For testing the various algorithms, we used
Stockfish \cite{StockfishEngine,StockfishGitHub}, the most popular {\em chess engine}
(i.e. chess-playing software), which is freely available for download. 
When playing at reduced strength (e.g., via the UCI-Elo or Skill Level
options), Stockfish introduces stochasticity into move selection.
The engine considers several candidate moves and applies a randomized
bias to the scores of slightly weaker moves, making suboptimal choices
more likely at lower strength levels \cite{StockfishFAQ}.
The strength of the engine can be set by applying its built-in command to determine its {\em Elo} level \cite{Elo1978}. For example, 1500 corresponds to a human player who had moved beyond beginner, and 3190 (the maximal level) is stronger than the world chess champion; different Elo levels produce notably different moves for the same position. 

\textbf{Empirical move randomness.}
To quantify the intrinsic randomness of the engine, and to ensure that the proposed algorithms are not learning repetitive patterns, we ran many engine--engine games and recorded both board positions and short sequences of moves. 
Specifically, we estimated the entropy of the position distribution after the 10th move, as well as the entropy of the sequences of moves 15-20, both over 100,000 games.
Let $p_i$ denote the empirical probability of outcome $i$ (a position or a move sequence) across repeated runs. 
We measured randomness using the Shannon entropy
\[
H = -\sum_i p_i \log_2 p_i ,
\]
which corresponds to an effective number $2^{H}$ of equally likely outcomes. 
The entropy of board positions after White's 10th move was $H = \text{16.607}$ bits (an effective number of $\text{99895}$ positions). For the sequence of moves from move 15 to move 20 we obtained an entropy of $H = \text{16.584}$ bits (an effective number of $\text{98262}$ sequences). Both numbers are very close to the number of games tested (100000), indicating 
a very large diversity of positions and move sequences across runs, consistent
with substantial variability in engine play.

\textbf{Basic engine features.}
The features of the engine we used are its choice of a move at a given position, as
well as its estimate of a position's strength for ${\cal \bf{W}}$, given by a
number between 0 and 1, which is referred to as "WDL"
(for Win-Draw-Lose). A value of $\alpha$ means that the engine
estimates the expected value of the game's result to be $\alpha$, and it is computed by $\frac{w + 0.5d}{1000}$, where $w$ resp. $d$ are the numbers of projected victories resp. draws if 1000 games are played from the current position.

The third party, in addition to ${\cal \bf{W}}$ and
${\cal \bf{B}}$, is the "helper" of the cheating party ${\cal \bf{W}}$.
This helper, denoted 
${\cal \bf{C}}$,
is assumed to be a player of championship strength, who can intervene and advise ${\cal \bf{W}}$ how to move. This
mimics cheating in the real world, where players use a chess engine tuned to a very high level, and play the moves suggested by it.
In all our experiments, 
the Elo levels of ${\cal \bf{W}}$ and ${\cal \bf{B}}$ were set to 1500,
and that of ${\cal \bf{C}}$ to 3190. These values are used only for the
experimental evaluation; the proposed methodology is independent of the
specific Elo levels and can be applied to other strength settings.

\subsection{Modifying the WDL value}
\label{section:isotonic}
\begin{figure}[t]
  \centering
  \includegraphics[width=0.49\linewidth]{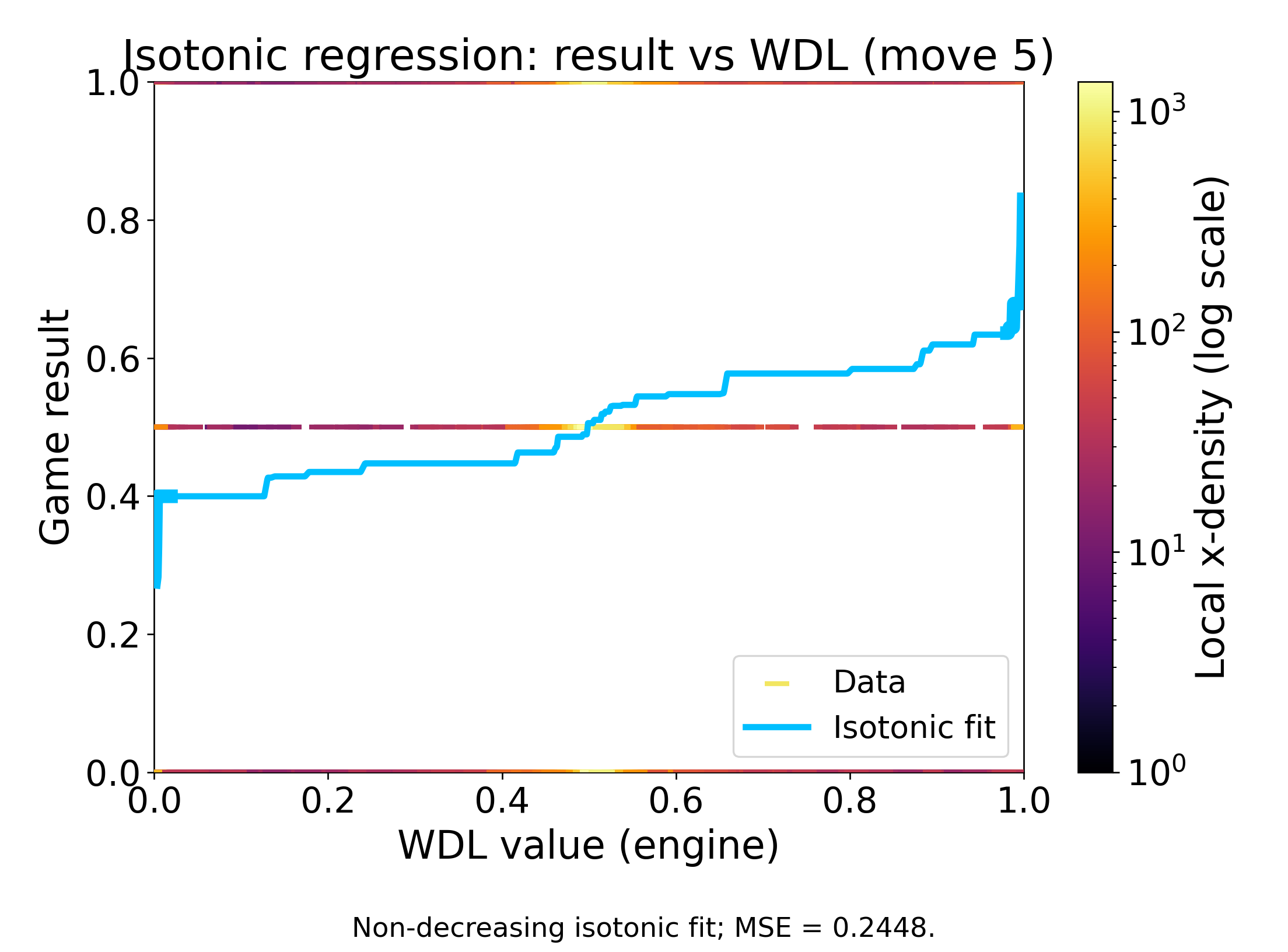}\hfill
  \includegraphics[width=0.49\linewidth]{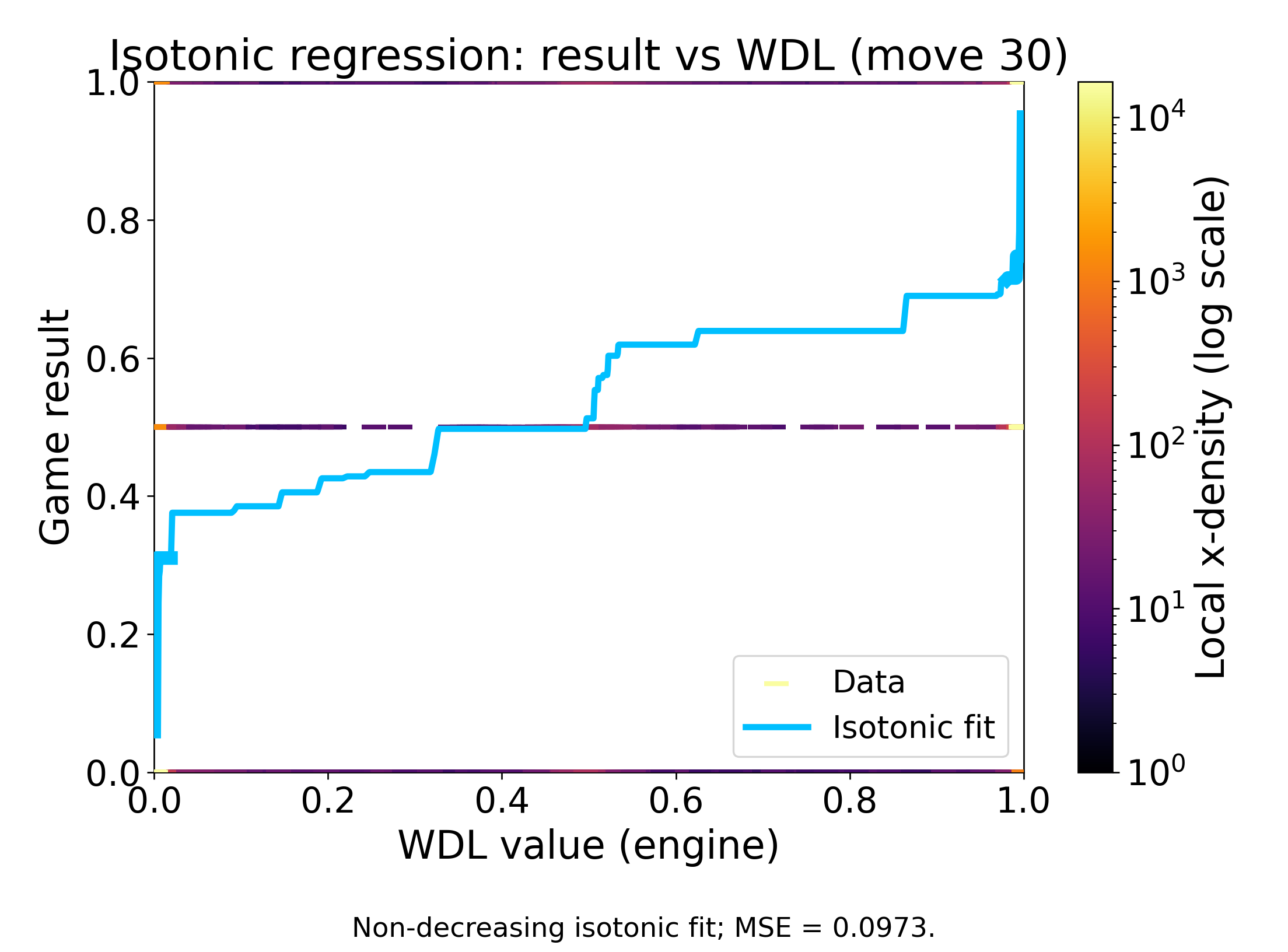}
  \caption{Isotonic regression: calibrated game result vs.\ engine WDL at move 5 (left) and move 30 (right). The wider effective range of the 30 move fit reflects the fact that the outcome is easier to predict when the game is in a more advanced stage.}
  \label{fig:isotonic}
\end{figure}

The WDL value provided by the Stockfish engine is crucial in deciding when to intervene, as it measures the quality of 
${\cal \bf{W}}$'s position; thus, a major factor in the decision is the difference between the WDL value after
the projected move of ${\cal \bf{W}}$, and the move offered by ${\cal \bf{C}}$. Our
experiments proved that substantially better results are
achieved when using a modified value of WDL, obtained by
fitting a function to empirical data defined, for every move
number $m$, by pairs $(\alpha_i,s_i)_m$, where $\alpha_i$ is
the engine WDL value at move $i$, and $s_i$ is game $i$'s
result. Since this function must be monotonically increasing
in $\alpha$,
we tried both {\em isotonic fit}, defined by the monotonic
function with the best MSE fit, which is known to be a step function \cite{deLeeuw2009}, as well as monotonically increasing neural network regression \cite{Sill1997}. The isotonic fit yielded lower MSE and better downstream results; hence we use it throughout all the experiments.
Results are presented in Fig.~\ref{fig:isotonic}.

\section{Basic Algorithms}
An {\em intervention} (or cheat) in the game is defined by ${\cal \bf{C}}$ -- the strong player, modeled by a maximal Elo engine -- watching as the game between ${\cal \bf{W}}$ and 
${\cal \bf{B}}$
(weaker players, modeled by an engine with lower Elo) progresses, and up to a fixed number of times $n$, tells ${\cal \bf{W}}$ how to move. Given $n$, we study various intervention algorithms, and check how they affect the game's average score. In Appendix A we provide an example of how ${\cal \bf{C}}$ improves over a move 
chosen by ${\cal \bf{W}}$.

All reported average scores, for the various algorithms in the paper, are averaged over $10^6$ games; the standard error of each reported mean is below 0.001.

We start with a naive algorithm, where ${\cal \bf{C}}$ chooses up to $n$ random moves and intervenes in them ("up to", as the game may end before the "intervention budget" $n$ is exhausted). This naive approach yielded a relatively modest improvement, presented in Table \ref{table:random}. 

\begin{table}[!htb]
  \centering
  \caption{Results for naive (random) intervention.}
  \begin{tabular}{@{}cc@{}}
    \toprule
    \textbf{Intervention budget} & \textbf{Average score} \\
    \midrule
    1 & 0.512 \\
    2 & 0.533 \\
    3 & 0.553 \\
    4 & 0.570 \\
    5 & 0.585 \\
    \bottomrule
      \label{table:random}
  \end{tabular}
\end{table}
\subsection{Fixed thresholds}
\label{section:fixed}
We next describe a simple threshold-based intervention algorithm.
Given a budget of $n$ interventions, then for every set of  $n$ 
positive thresholds $T_1 \ldots T_n$, ${\cal \bf{C}}$'s intervention policy 
is defined as follows: when the game starts, after each move by ${\cal \bf{B}}$, ${\cal \bf{C}}$ simulates a few moves with
the Elo level of ${\cal \bf{W}}$, computes the average WDL value after these moves are simulated, and compares that to the average WDL value following its moves (the number used for computing both averages was 10, and results were similar for a higher number). If the average WDL following 
${\cal \bf{C}}$'s moves is larger by at least $T_1$ than the
average following ${\cal \bf{W}}$'s moves, then 
${\cal \bf{C}}$ intervenes, advising ${\cal \bf{W}}$
how to move (by choosing its own move with the maximal WDL
score). Then, the game continues, and the same process is repeated with $T_2$. This  continues until the game ends, with ${\cal \bf{C}}$ allowed to use at most $n$ interventions.

In order to obtain an accurate estimate of the best thresholds
$T_i$, we ran many games with different thresholds, computed average scores over $10^6$ games for each set of $T_i$'s, and used Bayesian optimization \cite{BOSnoek2012PracticalBO} to find the optimal parameters. The results for 
$n=1,2,3$ are presented in Table~\ref{tab:thresholds}. The
 running time for the three-intervention optimization was very high (over a month, on a 
32-core server). In Section~\ref{section:stochastic}, we 
describe an approximation method for optimization of hyper-parameters such as these
thresholds, which does not use the chess engine at all, but relies
on a stochastic model for the WDL sequences, and applies it to find optimal thresholds.

\begin{table}[!htb]
\centering
\begin{tabular}{c l c}
\toprule
$n$ & Thresholds & Average Score \\
\midrule
1 & $T_1 = 0.205$ & 0.656 \\
2 & $T_1 = 0.141,\ T_2 = 0.199$ & 0.762 \\
3 & $T_1 = 0.133,\ T_2 = 0.147,\ T_3 = 0.193$ & 0.838 \\
\bottomrule
\end{tabular}
\caption{Thresholds and average scores for $n=1,2,3$.}
\label{tab:thresholds}
\end{table}
Note that even one intervention of this type yields substantially better performance than five random ones. The threshold for 
the $i$-th intervention is lower than for the ($i$+1)-th, meaning earlier interventions
are  more readily triggered. As expected,
the average scores follow a "diminishing returns" pattern (improvements for an additional intervention drop when the number of intervention increases). 
\section{Bellman-Based Intervention Policies}
\label{section:bellman}
Here we follow a more principled algorithm for deciding when
to intervene.
Crucially, the value of an intervention is not determined by the immediate
evaluation change at the intervention move, but by its effect on the
\emph{final game result} under subsequent weak--weak play.

The proposed policy is derived from the
Bellman optimality principle for a one-intervention stopping
problem: the algorithm compares the immediate uplift of intervening
now with the expected value of waiting for future opportunities
\cite{bellman1957}.
\subsection{Counterfactual Evaluation Principle}
At a position $s_t$, where $s$ denotes the game state and $t$ the move
number, we compare two counterfactual
continuations:

\begin{itemize}
\item \textbf{Weak branch:} play a weak move at $s_t$, then continue weak--weak
      until the end of the game;
\item \textbf{Strong branch:} play a strong move at $s_t$, then continue
      weak--weak until the end of the game.
\end{itemize}

The effect of an intervention is defined as the difference in the
expected final score between these two branches. In practice, we estimate
this quantity from game data; next, the implementation for one and two optimal interventions is described.
\subsection{Bellman-1: Optimal Single Intervention}
\label{sec:bellman1}
Let $U(s_t)$ denote the estimated counterfactual gain obtained by intervening
at position $s_t$, as defined by the difference between the strong and weak
branches described above.

Let $V_\infty(s_t)$ denote the maximal expected counterfactual gain attainable
by \emph{waiting}, i.e.\ by not intervening at $s_t$ and instead intervening
optimally at some later position.

The Bellman-1 decision rule is:
\[
\text{intervene at } s_t
\quad \text{if and only if} \quad
U(s_t) \;\ge\; V_\infty(s_t) + \theta_1,
\]
where $\theta_1$ is a tunable margin parameter (optimized by a grid
search from $-0.1$ to 0.3, in steps of 0.01), and $V_\infty(s_t)$ denotes
the expected maximal future uplift if we do not intervene at $s_t$.
It is estimated by regression from game logs: for each
position $s_t$, the regression target is the maximal uplift observed at
later moves, and the predictor uses the current state
features. The regression (here and in the rest of the paper) uses a
random forest model trained on 100{,}000 stored games. The features
include the current move number, the calibrated strong/weak WDL values
over the last five moves, and the corresponding counterfactual values
for the next move. "Looking back" more than five moves made no difference;
and for moves 1--5, no intervention scheme we tried selects them as
intervention points, since the game is still in its early stage.

In similar fashion, we define the Bellman-1 value function
\[
V_1(s_t) = \max\{\, U(s_t),\; V_\infty(s_t) \,\},
\]
which represents the optimal expected uplift when one intervention
remains and the current position is $s_t$.
\subsection{Bellman-2: Optimal Double Intervention}
\label{sec:bellman2}
The Bellman-2 policy extends the above reasoning to the case of at most two interventions.

After a first intervention at $s_t$, the game transitions to a position
$s_{t+1}^{\text{strong}}$, from which a second intervention may be optimally chosen.

Let $V_1(s)$ denote the Bellman-1 value function, i.e.\ the optimal single-intervention
value starting from position $s$.

Note: after the first intervention, the game may reach positions that are atypically
strong for weak-engine play. Thus, the value function $V_1(s)$ is applied
to states that are somewhat outside the distribution on which it was trained.
This introduces an approximation; however, it works
well in practice.

The counterfactual value of intervening at $s_t$, while
leaving one intervention for later, is
defined as
\[
U_2(s_t)
\;=\;
\mathbb{E}\!\left[ V_1(s_{t+1}^{\text{strong}}) \right]
\;-\;
\mathbb{E}\!\left[ V_1(s_{t+1}^{\text{weak}}) \right].
\]

Here $U_2(s_t)$ represents the marginal uplift obtained by intervening
at move $t$, relative to continuing with the weak move. The total value
of intervening is therefore
\[
V_1(s_{t+1}^{\text{weak}}) + U_2(s_t)
= \mathbb{E}[V_1(s_{t+1}^{\text{strong}})].
\]

As in the single-intervention case, this quantity is compared against the value
of waiting:
\[
\text{intervene at } s_t
\quad \text{if and only if} \quad
U_2(s_t) \;\ge\; V_\infty^{(2)}(s_t) + \theta_2,
\]
where $V_\infty^{(2)}(s_t)$ denotes the maximal expected future value
attainable by waiting at $s_t$ and using the two remaining interventions
optimally at later positions, and $\theta_2$ is optimized by grid search
in the same manner as $\theta_1$ (Section~\ref{sec:bellman1}).
\subsection{Discussion}
The policies in Section \ref{sec:bellman1}, \ref{sec:bellman2} evaluate intervention decisions directly through their effect on
the \emph{final game outcome}. Intermediate evaluation changes at the
intervention move are used only as predictors of this final outcome, rather
than as optimization targets themselves. This yielded better results than the "local" approach described in Section \ref{section:fixed}.

Results are presented in Table~\ref{tab:bellman}. 

\begin{table}[!htb]
\centering
\begin{tabular}{l c}
\toprule
\textbf{Policy} & \textbf{Average Score} \\
\midrule
No intervention & 0.515 \\
Bellman-1 (one intervention) & 0.705 \\
Bellman-2 (two interventions) & 0.822 \\
\bottomrule
\end{tabular}
\caption{Average scores for the Bellman-based policies.}
\label{tab:bellman}
\end{table}
\subsection{Hindsight}
We also tested a ``hindsight'' version of the Bellman-1 policy: after the decision to intervene, the game continues, but in parallel, we
open another ``branch'' in which ${\cal \bf{C}}$ waits
and advises to intervene the next time at which the condition for
intervention holds. The result is defined as the maximum of the final game results of both branches; in lay terms, ${\cal \bf{C}}$ ``sees into the future'' and chooses the best option among the two. The hindsight improvement is substantial (0.743 vs.\ 0.705), which
suggests that future work may optimize over the decision whether to 
intervene at the first time the condition is met, or wait for the next opportunity.

\section{Using a stochastic, engine-free model}
\label{section:stochastic}
We now present an approximate algorithm for tuning
hyperparameters (for example, the thresholds in Section \ref{section:fixed}), not by running games on the chess engine, 
but simulating interventions on WDL sequences learned from games.

\paragraph{Motivation}
As in Section~\ref{section:fixed}, we wish to choose when to replace a weak move by a strong one during a game, using up to \(n\) interventions with thresholds \(T_1 \ldots T_n\) (but the idea presented in this section
applies to any choice of hyperparameters, for example $\theta_1,\theta_2$ in the
Bellman-1 and Bellman-2 algorithms).

Evaluating the uplift for every candidate \(T_1 \ldots T_n\) is prohibitively slow, and suffers from the "curse of dimensionality": it took weeks on a strong server to optimally tune three thresholds
using Bayesian optimization, and running time steeply increases with the
number of variables to optimize over.

Instead, we use an \emph{engine-free} simulator built from logs: (i) per--move-number ($t$) banks of WDL scores \((p_w(t),p_s(t))\) from games without intervention (where $p_w(t)$ resp. $p_s(t)$ refer to weak resp. strong moves),
and (ii) an empirical "uplift"
\(\Delta(t,d)\) learned from games with randomized single interventions, where \(d=\max\{0,\,p_s-p_w\}\) is the strong–weak gap. We used 100,000 games of each type.

This approach provides a fast estimate of expected score for any 
\(T_1 \ldots T_n\), enabling rapid Bayesian optimization with a clear diagnostics (how often and when interventions are triggered), while remaining faithful to observed play.
\subsection*{Engine-free threshold simulation}
\paragraph{Setup and notation}
We entirely forgo running the chess engine, and strip games
down to the sequence of calibrated outcomes, depending on
whether a weak or strong move is currently taken.
For each position after ${\cal \bf{W}}$'s $t$-th move ($t=1,2,\dots,H$, where $H$ is a threshold over the number of moves, chosen to be very large -- e.g. 200), we assume 
we have expected scores  under a weak and a strong move,
denoted
\[
p_w(t),\quad p_s(t) \in [0,1].
\]
These quantities are estimated in advance from engine-game logs,
using the same calibrated expected-score framework described earlier (Section \ref{sec:bellman1}).

Define the (nonnegative) strong--weak gap
\[
d(t) \;=\; \max\{\,0,\; p_s(t) - p_w(t)\,\} 
\]
and let the final game result be \(Y\in\{0,\tfrac12,1\}\).

\paragraph{Data}
We use two logs:
(i) a \emph{no-intervention} set \(\mathcal D_0\) containing, for many games, the
sequence \(\{(t, p_w(t), p_s(t))\}\) and the final score \(Y\);
(ii) a \emph{random single-intervention} set \(\mathcal D_I\) in which one move
number \(t^{\!*}\) per game is chosen at random and the \emph{strong} move is played there, with final score \(Y\) recorded. For fast replay we build a bank of pairs,
\[
\mathcal B_t \;=\; \{\, (p_w, p_s) \text{ observed at move } t \text{ in } \mathcal D_0 \,\},
\]
\paragraph{The "uplift" function \(\Delta(t,d)\)}
We model the expected final score improvement when replacing the weak move by
the strong move at move \(t\), given the gap \(d\):
\[
\Delta(t,d) \;\approx\; \mu_1(t,d)\;-\;\mu_0(t,d),
\]
where \(\mu_1(t,d)=\mathbb E[Y\mid I=1,t,d]\) is estimated from
\(\mathcal D_I\) (games intervened at \(t\)) and \(\mu_0(t,d)=\mathbb E[Y\mid I=0,t,d]\)
from \(\mathcal D_0\). Each move number $t$ is assigned a bin,
 \(b(t)\),
and in each bin we fit \(\mu_1^b(d)\) and \(\mu_0^b(d)\) as monotone in \(d\) curves,
then set \(\Delta(t,d)=\mu_1^{b(t)}(d)-\mu_0^{b(t)}(d)\). We pre-compute a look-up
table over \(d\in[0,1]\) for each move bin.
\paragraph{Threshold policy and simulator}
We simulate games without engine calls by sampling calibrated pairs
\((p_w,p_s)\) from the no-intervention bank \(\mathcal B_t\) at each move
number \(t\). Let \(d(t)=\max\{0,p_s-p_w\}\) denote the strong--weak gap.
When \(d(t)\) exceeds the next threshold \(T_k\), the \(k\)-th intervention
is used at move \(t\). Each intervention contributes an uplift
\(\Delta(t,d)\) to the expected score. The simulation proceeds as follows.

\begin{enumerate}
\item Initialize \(v \leftarrow \bar y_0\), where \(\bar y_0\) is the empirical
      average final score in the no-intervention dataset \(\mathcal D_0\).
      Thus, the simulator estimates policy performance relative to the
      population no-intervention baseline, rather than to a game-specific
      baseline for the sampled trajectory.
\item For moves \(t=1,\dots,H\): sample \((p_w,p_s)\sim \mathcal B_t\) and set
      \(d=\max\{0,p_s-p_w\}\). Let $k$ denote the number of interventions used so far. If \(k < n\) and \(d\ge T_k\), record
      \((k,t,d)\) and increment \(k\).
\item For each recorded intervention \((k,t,d)\), update
      \[
      v \leftarrow \mathrm{clip}\!\bigl(v + \Delta(t,d),\,0,1\bigr).
      \]
\item Return \(v\) as the simulated expected final score.
\end{enumerate}

\paragraph{Monte Carlo optimization for optimal thresholds}
We repeat the simulator for \(N\) independent runs and report
\[
\mathrm{AvgScore}(T_1 \ldots T_n) \;=\; \frac{1}{N}\sum_{i=1}^{N} v^{(i)}.
\]

As in Section~\ref{section:fixed}, we used Bayesian optimization over the simulated average score, and optimized for four thresholds.
\paragraph{Results}
We ran the engine-free optimization for three and four thresholds. Running time was much lower than running
directly over games; for four thresholds, even when averaging $10^6$ runs per a single candidate $T_1,T_2,T_3,T_4$, the
optimization took about 8 hours. The optimum was obtained
at $0.15437,0.17808,0.19909,0.24803$ (note that, as for three
thresholds in Section \ref{section:fixed}, the sequence of values is increasing, reflecting a more
stringent intervention criterion as they are carried out). The
average score was 0.907 (note -- this score was obtained by
running engine games, at the optimal point found for the 
approximation non-engine games).

In order to compare to the "ground truth" (running on engine
games), we ran the engine-free optimization on three thresholds,
which yielded a nearly identical result to the one reported in
Section~\ref{section:fixed} (0.835 vs.\ 0.838), reflecting
good correspondence between the "engine-free" approximation 
and the optimal engine result.
\section{Relevance to human-vs-human cheating}
\label{section:human-human}
Our experiments use engine-vs-engine play. To estimate relevance to
human play, we compare \emph{conversion rates}: given that White
reaches a position with calibrated expected score $\alpha$ at
move~$n$, what is the average final score?

Table~\ref{tab:conversion} compares engine self-play (Elo 1500) to
human games from Lichess.org (Elo 1500--1520). Engines convert
advantageous positions more reliably than humans, who are more prone
to calculation errors and time pressure. The average gap is
$\delta\approx 0.08$, somewhat larger at later moves and higher advantages, where precise calculation matters most. 
Thus our engine-based scores should be interpreted as an approximate
upper bound. Subtracting $\delta$ yields conservative
human-realistic estimates (e.g., Bellman-1 score of
$0.71 - 0.08 = 0.63$, vs.\ $0.51$ with no cheats --- still a
substantial gain).

\begin{table}[!htb]
\centering
\caption{Conversion rates. Average final scores (rounded to two decimal places), after 
 ${\cal \bf{W}}$ reaches calibrated advantage $\alpha$ at move $n$.
Human play data taken from Lichess (Elo 1500--1520). Note that both humans and engine are better at converting advantages obtained in later stages of the game.}
\label{tab:conversion}
\begin{tabular}{@{}c c c c c@{}}
\toprule
& \multicolumn{2}{c}{Move 15} & \multicolumn{2}{c}{Move 30} \\
\cmidrule(lr){2-3} \cmidrule(lr){4-5}
$\alpha$ & Engine & Human & Engine & Human \\
\midrule
0.55 &  0.53      &  0.51      &  0.57      &   0.52     \\
0.60 &   0.58     &    0.52    &    0.61    &    0.53    \\
0.65 &    0.62    &    0.55    &    0.64    &    0.57    \\
0.70 &    0.64    &    0.56    &     0.72   &     0.61   \\
0.75 &     0.67   &    0.58    &    0.81    &    0.64    \\
\bottomrule
\end{tabular}
\end{table}

\section{Conclusions and future work}
We studied the effectiveness of limited cheating in chess, where a strong engine advises a weaker one in a small number of moves. Using threshold-based policies optimized via Bayesian optimization, a single well-chosen intervention raises the average score from 0.51 to 0.66; using a Bellman-based counterfactual framework, a single intervention achieves 0.71, and two interventions achieve 0.822. The Bellman-based approach provides a principled basis for deciding when to intervene, while an engine-free stochastic simulator enables fast hyperparameter tuning that very closely matches engine-based results. These findings quantify the substantial advantage that even minimal, carefully timed engine assistance provides -- information that is valuable for understanding and ultimately detecting cheating in competitive chess.

Future work will focus on improving the current results. In games where white leads from the beginning, or in which White achieves a winning position after less than $n$ cheats, often less than the allotted  $n$ interventions are used. For example, if $n=3$, only 44\% of the games use three interventions. Another direction is to allot an "average cheat budget" over many games, and use more when necessary, at the expense of games in which ${\cal \bf{W}}$ requires less interventions.

Another important line of research is cheating detection. The results presented here suggest that even cheating in a relatively small number of moves can significantly increase the average score. This raises the question of how to deal with such cheating tactics as those studied here.

Lastly, we plan to explore how the algorithms presented here apply in human vs. human games, and obtain more complete results than those briefly presented in Section \ref{section:human-human}.

\bibliographystyle{plainnat}
\bibliography{chess}

\appendices
\section{Example of an intervention}
Fig.~\ref{wsm} shows how the strong engine finds a better move (blue arrow) than the weak engine (red arrow) in the same position, correctly anticipating Black doubling rooks on the A-file.
\bigskip
\begin{center}
  \includegraphics[width=0.4\columnwidth]{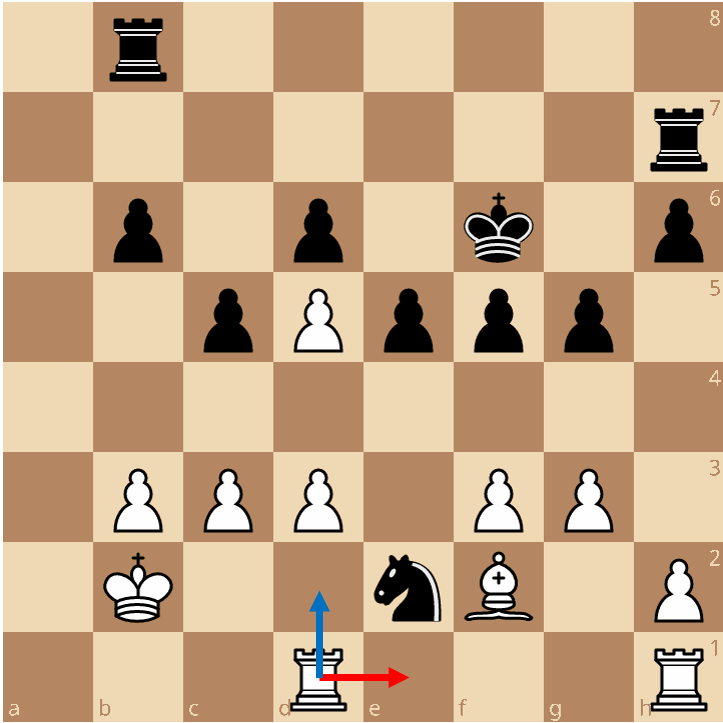}
  \captionof{figure}{"Weak" (red arrow) vs. "strong" (blue arrow) moves.}
  \label{wsm}
\end{center}
\end{document}